\documentclass[sigconf,table,bookmarksnumbered,unicode,hidelinks]{acmart}
\AtBeginDocument{%
  }

\usepackage{array}
\usepackage{multirow}
\usepackage{makecell}
\usepackage{booktabs}
\usepackage{algorithm}
\usepackage{algorithmic}
\usepackage[switch]{lineno}
\usepackage{caption}
\usepackage{hhline}
\usepackage{pifont}  
\usepackage{graphicx}

\newcommand{\cmark}{\ding{51}}

\usepackage{balance}
\copyrightyear{2025}
\acmYear{2025}
\setcopyright{acmlicensed}\acmConference[MM '25]{Proceedings of the 33rd ACM International Conference on Multimedia}{October 27--31, 2025}{Dublin, Ireland}
\acmBooktitle{Proceedings of the 33rd ACM International Conference on Multimedia (MM '25), October 27--31, 2025, Dublin, Ireland}
\acmDOI{10.1145/3746027.3755876}
\acmISBN{979-8-4007-2035-2/2025/10}

\begin{document}

\title{DepthGait: Multi-Scale Cross-Level Feature Fusion of RGB-Derived Depth and Silhouette Sequences for Robust Gait Recognition
}

\author{Xinzhu Li}
\orcid{0009-0009-8161-2626}
\affiliation{%
  \institution{Sun Yat-sen University}
  \city{Zhuhai}
  \country{China}
}

\author{Juepeng Zheng}
\affiliation{%
  \institution{Sun Yat-sen University}
  \city{Zhuhai}
  \country{China}}

\author{Yikun Chen}
\affiliation{%
  \institution{Guangdong Zhiyun Urban Construction Technology Co., Ltd.}
  \city{Zhuhai}
  \country{China}
}

\author{Xudong Mao}
\affiliation{%
 \institution{Sun Yat-sen University}
 \city{Zhuhai}
 \country{China}}

\author{Guanghui Yue}
\affiliation{%
  \institution{Shenzhen University}
  \city{Shenzhen}
  \country{China}}

\author{Wei Zhou}
\affiliation{%
  \institution{Cardiff University}
  \city{Cardiff}
  \country{UK}}

\author{Chenlei Lv}
\affiliation{%
  \institution{Shenzhen University}
  \city{Shenzhen}
  \country{China}}

\author{Ruomei Wang}
\affiliation{%
  \institution{Sun Yat-sen University}
  \city{Zhuhai}
  \country{China}}

\author{Fan Zhou}
\affiliation{%
  \institution{Sun Yat-sen University}
  \city{Guangzhou}
  \country{China}}

\author{Baoquan Zhao}
\authornote{Corresponding author. E-mail: zhaobaoquan@mail.sysu.edu.cn}
\affiliation{%
 \institution{Sun Yat-sen University}
 \city{Zhuhai}
 \country{China}}

\renewcommand{\shortauthors}{Xinzhu Li et al.}

\begin{abstract}
Robust gait recognition requires highly discriminative representations, which are closely tied to input modalities. While binary silhouettes and skeletons have dominated recent literature, these 2D representations fall short of capturing sufficient cues that can be exploited to handle viewpoint variations, and capture finer and meaningful details of gait. In this paper, we introduce a novel framework, termed DepthGait, that incorporates RGB-derived depth maps and silhouettes for enhanced gait recognition. Specifically, apart from the 2D silhouette representation of the human body, the proposed pipeline explicitly estimates depth maps from a given RGB image sequence and uses them as a new modality to capture discriminative features inherent in human locomotion. In addition, a novel multi-scale and cross-level fusion scheme has also been developed to bridge the modality gap between depth maps and silhouettes. Extensive experiments on standard benchmarks demonstrate that the proposed DepthGait achieves state-of-the-art performance compared to peer methods and attains an impressive mean rank-1 accuracy on the challenging datasets.
\end{abstract}



\maketitle

\section{Introduction}
Human gait recognition is an appealing technology that identifies individuals based on their unique walking patterns, which encompass various aspects of gait, including stride length, walking speed, body posture, and limb movements. In contrast to traditional biometrics such as face, fingerprint, and iris, this non-invasive, long-distance, and potentially more difficult-to-forge method has garnered significant attention in the evolving landscape of biometric identification due to its wide spectrum of applications in various sectors such as security, surveillance, health, forensic science, and beyond~\cite{sepas2022deep,filipi2022gait, shen2024comprehensive, sheng2023data, Cosma_2024_WACV}. Despite the efforts made so far, gait recognition remains a challenging task due to several factors. Variability in walking conditions, such as changes in footwear, clothing, and carrying objects, can significantly alter an individual's gait, making consistent recognition difficult. Additionally, environmental factors like lighting, camera angles, and surface conditions can affect the accuracy of gait analysis. Furthermore, distinguishing subtle differences in gait patterns across different individuals while accounting for natural variations within the same person over time adds another layer of complexity. These challenges necessitate ongoing research to enhance the robustness and reliability of gait recognition.
\begin{figure}[!t]
\centering
\includegraphics[width=0.49\textwidth]{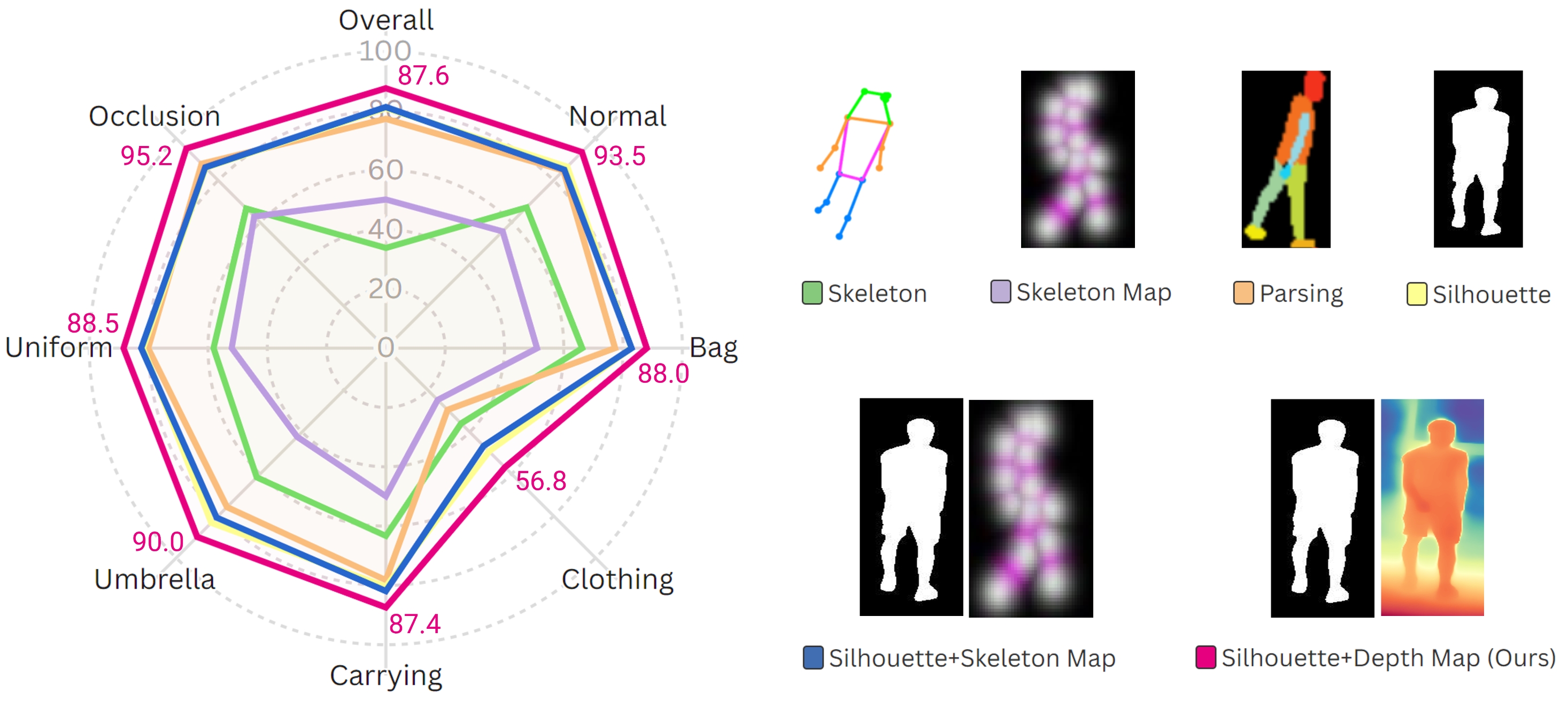}
\caption{Quantitative evaluation on the SUSTech1K dataset of different gait recognition approaches. The radar chart (left) illustrates the overall performance and seven challenging test protocols.  The proposed method (shown in pink) consistently outperforms the baseline approaches, including Skeleton, Skeleton Map, Parsing, Silhouette, and Silhouette+Skeleton Map.}
\label{fig1}
\end{figure}


Traditionally, binary silhouette sequences and skeletons derived from RGB motion videos have dominated as the primary input representations for existing gait analysis pipelines. Silhouette sequences excel in distinguishing individuals by explicitly preserving appearance information, while skeletons retain the internal structural information of the human body, offering natural robustness to appearance changes. However, both modalities have inherent limitations. Silhouettes are susceptible to significant changes in external body shape due to variations in clothing, while skeletons, though effective in addressing clothing occlusion, completely disregard distinguishing body shape information, leading to suboptimal performance. Additionally, alternative gait representations such as parsing~\cite{10.1145/3581783.3612052} and skeleton map~\cite{fan2024skeletongait} have also shown potential. To address the limitations of monomodality, recent studies ~\cite{Cui_2023_CVPR,fan2024skeletongait} have explored the fusion of these two modalities, yielding increased performance (see Figure~\ref{fig1}).  Nevertheless, existing solutions still fall short of effectively addressing the complexities of real-world scenarios.

Inspired by the recent advancement of depth estimation foundation models like Depth Anything~\cite{depthanything}, we formulate a hypothesis that \textit{the use of depth information aligns more closely with how humans perceive and recognize gait patterns, potentially leading to more intuitive and effective gait recognition systems.} This could be reasonable for several reasons. Firstly, depth maps provide more explicit 3D geometric information about the human body and its movements, which is not directly available in silhouettes and skeletons. This additional dimension allows for a more comprehensive representation of gait dynamics. Secondly, depth maps allow for more accurate analysis of subtle motions and joint movements, which are crucial for capturing the unique characteristics of an individual's gait. Moreover, depth information can help in addressing the challenge of viewpoint variations, as it provides a more consistent representation of the body's structure and movement across different angles. 

In addition, when used in conjunction with other modality data, depth maps are likely to offer complementary information that can enhance the overall discriminative power of gait recognition systems. In this case, an effective feature fusion strategy is thus essential for integrating dual data modalities for robust gait recognition. It's tempting to think that we can leverage the aforementioned schemes developed for silhouette and skeleton feature fusion. Unfortunately,
existing multimodal fusion techniques either fail to capture and integrate the complex relationships between different modalities, or fall short of modeling fine-grained gait features required in gait recognition tasks.

To address aforementioned challenges, we introduce DepthGait, a novel dual-branch framework for robust gait recognition that, for the first time, utilizes both silhouette and depth map sequences derived from human motion RGB videos using advanced foundation models. To facilitate feature fusion of these two modalities, we also present a novel architecture designed for multi-scale and cross-level deep feature fusion of the two branches. Extensive experiments on widely used gait recognition benchmarks demonstrate the superiority of the proposed DepthGait over several state-of-the-art methods. The main contributions of this paper are summarized as follows:
\begin{itemize}
    \item We pioneer the use of RGB-derived depth maps in conjunction with silhouettes for gait recognition. Our findings demonstrate that it outperforms widely used silhouette and skeleton modality inputs, opening new avenues for future gait research.
    \item We feature a new network for effectively fusing depth map and silhouette sequences. Through multi-scale attention and cross-level fusion, we enhance inter-modality interaction, allowing for more fine-grained information integration between modalities.
    \item Experimental results demonstrate the superiority of the proposed DepthGait, achieving state-of-the-art performance on widely-used gait recognition benchmarks.
\end{itemize}

\begin{figure*}[!t]
\includegraphics[width=\textwidth]{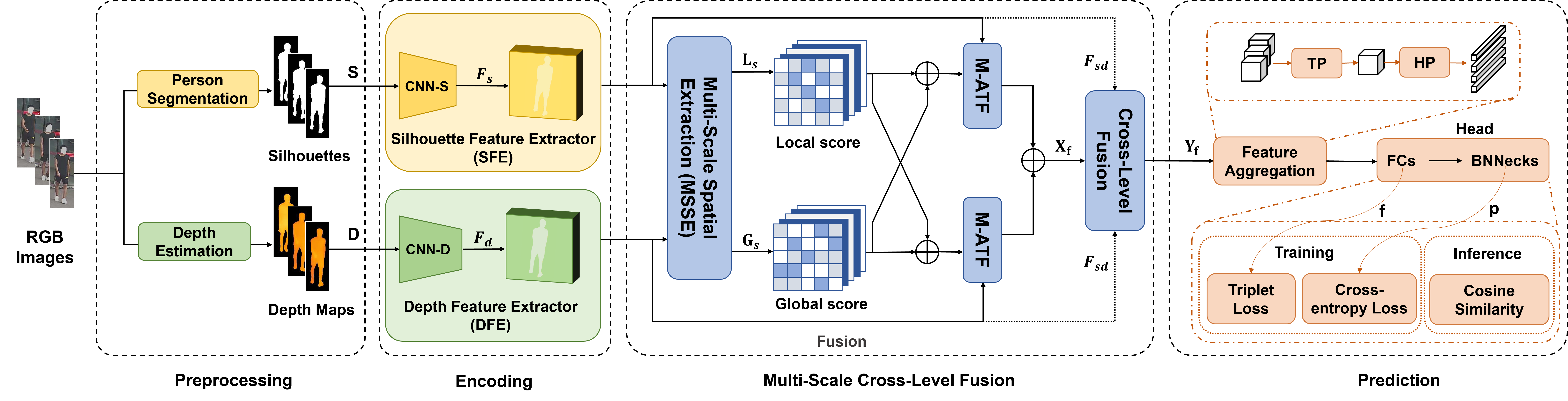} 
\caption{The architecture of the proposed DepthGait model for robust gait recognition.}
\label{fig2}
\end{figure*}
\section{Related Work}
 
\textbf{2D Gait Recognition.} Methods in this category primarily learn gait representations from binary silhouette images extracted from RGB sequences. This representation is well-suited for long-distance, low-resolution conditions, contributing to its popularity~\cite{shiraga2016geinet,huang20213d,hou2020gait,10.1145/3503161.3547897,peng2024glgait}. GaitSet~\cite{chao2019gaitset} pioneered treating gait sequences as sets, using CNNs to extract frame-level features independently from silhouettes before aggregating them into set-level features. GaitPart~\cite{fan2020gaitpart} models temporal dependencies in local silhouette details, providing distinct spatiotemporal representations for each body part. GaitGraph~\cite{teepe2021gaitgraph} combines graph and 2D convolutional layers, emphasizing structural features and spatiotemporal modeling of skeleton graphs.
Recently, DeepGaitV2~\cite{fan2024exploringdeepmodelspractical} emphasized deeper gait feature extraction networks, achieving state-of-the-art performance on popular benchmarks. Despite this progress, more discriminative gait representations are highly desired to tackle challenging scenarios effectively.

\textbf{3D Gait Recognition.} Unlike methods relying on 2D silhouettes or skeletal structures, approaches in this category~\cite{bouchrika2007model,ariyanto2011model,liao2017pose,han2024gait,li2025MSPoint} model gait patterns using 3D data representations, including 3D skeletons and parameter-based models like SMPL~\cite{loper2015smpl}. PoseGait~\cite{liao2020model} leverages 3D skeleton data and human prior knowledge to mitigate external variations, while GaitSG~\cite{yan2023gaitsg} utilizes SMPL models as input, modeling them as graphs with graph convolution techniques.
Beyond 3D gait data derived from RGB images, there is an emerging trend of using 3D point clouds captured with LiDAR sensors. LidarGait~\cite{shen2023lidargait} learns gait representations with enriched 3D geometry from sparse point clouds, demonstrating LiDAR's superiority over RGB cameras. Point cloud solutions offer advantages in high privacy environments compared to optical cameras.
In contrast, we propose the first approach to directly derive 3D gait information by leveraging recent monocular depth estimation foundation models. Compared to LiDAR-acquired depth data, our RGB-derived depth information offers several advantages: leveraging existing cost-effective camera infrastructure in security systems and enabling retrospective application to existing RGB video datasets for gait recognition research on historical data.

\textbf{Multi-modal Gait Recognition.} By combining modalities such as silhouette, skeleton, and 3D SMPL models, multimodal methods~\cite{castro2020multimodal,delgado2018end,hofmann2014tum} capture comprehensive and complementary information, addressing mono-modality limitations. SMPLGait~\cite{zheng2022gait} introduced 3D mesh into gait recognition, employing 3D spatial transformer networks (3D-STN) to normalize human postures across viewpoints. BiFusion~\cite{peng2024learning} integrates skeleton and silhouette information using a novel multi-scale gait graph (MSGG) network that hierarchically extracts gait patterns from raw skeleton data. SkeletonGait++~\cite{fan2024skeletongait} introduced skeleton maps, representing joint coordinates as heatmaps using a Gaussian approximation to enhance skeletal-silhouette integration.
Feature fusion plays a vital role in effective multimodal gait recognition. SMPLGait~\cite{zheng2022gait} employs straightforward element-wise operations, which may be insufficient for bridging modality gaps and extracting mutual information, resulting in limited performance enhancement from the 3D branch. While previous studies introduced attention mechanisms, most rely on simple fusion operations. SkeletonGait~\cite{fan2024skeletongait} applied basic attention fusion only during the second encoding stage, insufficient for capturing complex inter-modality relationships.
These deficiencies motivated us to develop a novel feature fusion solution to leverage the merits of our proposed multimodal gait recognition framework with a newly developed depth-based gait representation.

\section{Method}

\subsection{Methodology Overview}
As shown in Figure~\ref{fig2}, our architecture consists of four modules: preprocessing, encoding, multi-scale cross-level fusion, and prediction. The pipeline first derives silhouette and depth map sequences from sampled RGB video frames during preprocessing. These sequences are fed into corresponding deep feature extractors in the encoding module to generate feature maps. The multi-scale cross-level fusion module extracts multi-scale features and combines them with original feature maps to produce the final gait representation. Finally, the prediction module processes this representation using a gait recognition framework for training and inference.

\subsection{Preprocessing Module}
Two types of gait representations are derived from the original video footage at the preprocessing stage.

\textbf{Silhouette Gait Representation.} Since well-processed silhouettes are generally available in existing widely-used gait recognition datasets, we don't reiterate the segmentation process in detail here. The silhouette sequence is represented as $S$, with dimensions $C_s^1 \times T_s \times H_s^1 \times W_s^1$. Here, $C_s^1$ represents the number of channels, $T_s$ denotes the length of the silhouette sequence, and $(H_s^1, W_s^1)$ corresponds to the image dimensions of each frame. 

\begin{figure}[!t]
\centering
\includegraphics[width=0.46\textwidth]{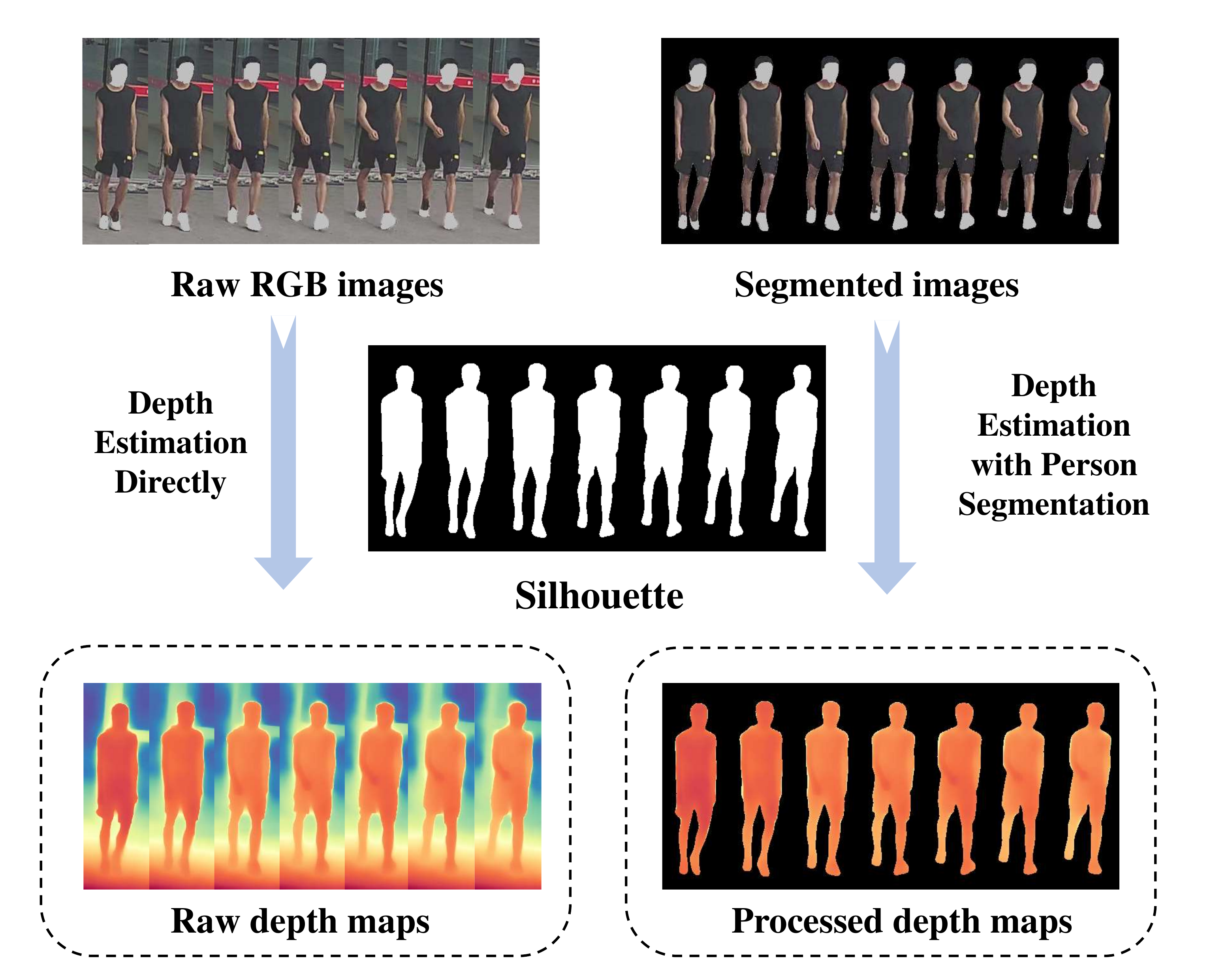}
\caption{Two ways of depth map acquisition. Depth estimation can be performed on either the raw RGB images or the segmented human images.}
\label{fig3}
\end{figure}

\textbf{Depth Map Gait Representation.} 
To capture 3D information of the human body, we employ the Depth Anything foundation model~\cite{depthanything} to estimate the depth maps of RGB images. As depicted in Figure~\ref{fig3},  depth estimation can be performed either on the raw RGB images or the human segmentation images. For the first case, we can use the silhouette as a mask to yield a depth map with the region of interest only. 
To facilitate subsequent feature extraction, the resultant depth map undergoes conversion to disparity space, followed by normalization. 
The depth value $d$ is transformed into its corresponding disparity $q = \frac{1}{d}$ using the inverse relationship. The normalized disparity values $q'$ within the range [0, 1] are calculated by:
$
q' = \frac{q - q_{min}}{q_{max} - q_{min}}.
$
This transformation and normalization process yields a more standardized feature map.

\begin{figure}[!t]
\centering
\includegraphics[width=0.46\textwidth]{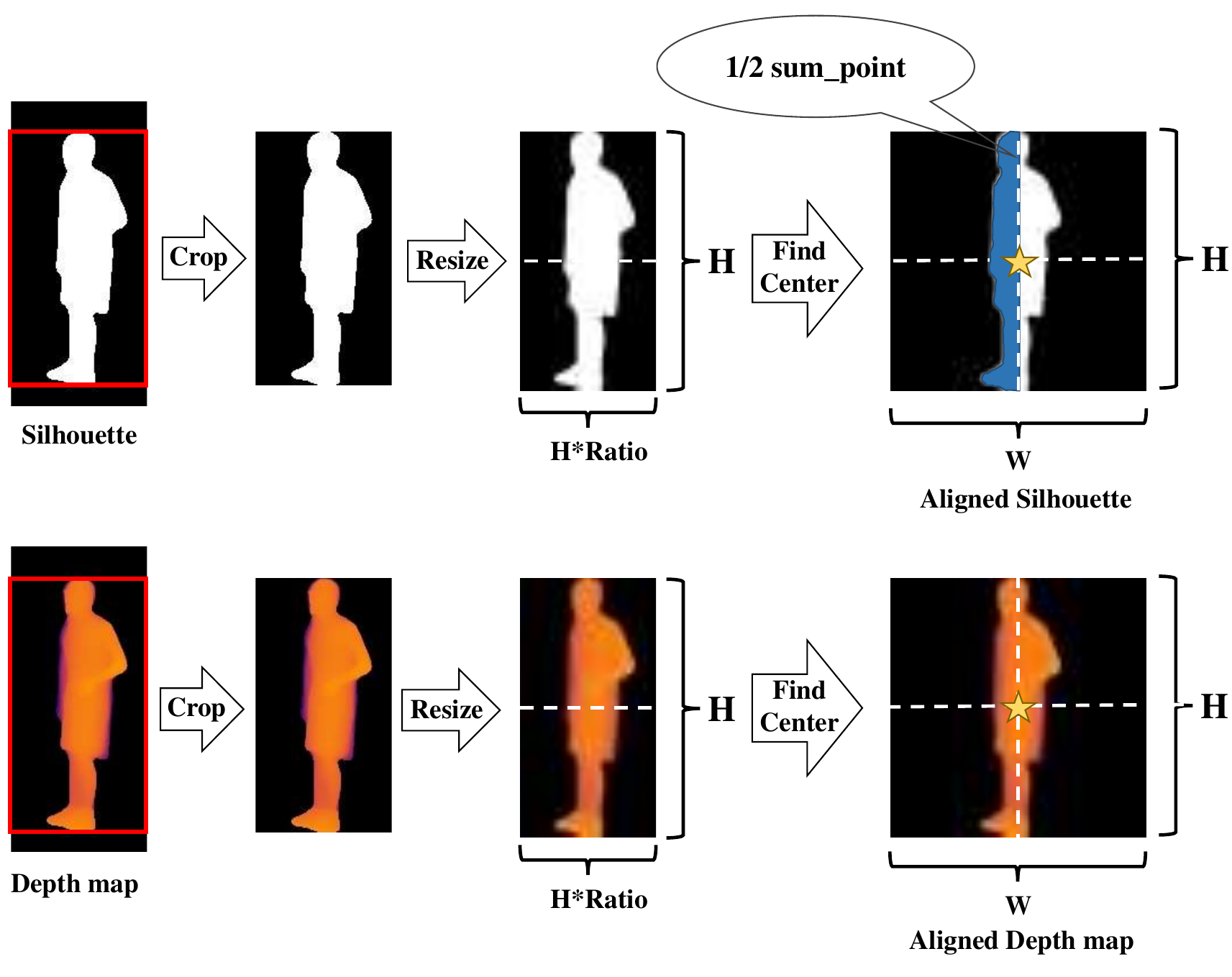} 
\caption{Normalization of silhouettes and depth maps.}
\label{figure4}
\end{figure}

To keep informative gait-related information in the obtained silhouettes and depth maps, we perform additional cropping and normalization as shown in Figure~\ref{figure4}. Initially, we identify the top and bottom positions of non-zero elements in both images, and crop the images to isolate the human figure. Following the standard practice in gait recognition, we set the silhouette input height to 64 pixels and adjust the widths of both silhouettes and depth maps accordingly, maintaining the original aspect ratio of the human body. This process ensures that the horizontal center is positioned at half of the image height, providing a consistent frame of reference across all samples.

For vertical center alignment, we opt to use the vertical center of the silhouette image as a reference for both silhouette and depth images, as it offers more reliable identification. To determine this central axis position $x_{center}$, we first calculate the total number of white pixels in the image, then compute the cumulative number of white pixels in each column from left to right. By traversing this cumulative pixel array, we identify the position where the cumulative pixel count exceeds half of the total pixel count, designating this as the central axis of the image. 
After determining the central axis, we proceed to calculate the left and right boundaries for cropping. This process takes into account the position of the central axis and ensures that the resulting cropped image maintains a standardized width $w$ (set to 64 pixels). If the calculated boundaries extend beyond the original image dimensions, we apply zero padding to both sides of the image. 
Once these boundaries are established, we apply the same cropping parameters to the depth image, maintaining alignment between the silhouette and depth representations. 
We represent the depth map sequence as $D$, with dimensions $C_d^1 \times T_d \times H_d^1 \times W_d^1$, where $C_d^1$, $T_d$, and $(H_d^1, W_d^1)$ represent the number of channels, sequence length, and frame dimensions for the depth maps, respectively.

\subsection{Encoding Module}
The resultant silhouette sequence $S$ and depth map sequence $D$ are then fed into the silhouette feature extractor (SFE) and depth feature extractor (DFE), respectively, in the encoding module. Given the impressive performance across various gait datasets, we employ the feature extractor introduced in DeepGaitV2~\cite{fan2024exploringdeepmodelspractical} as our encoder to harvest feature maps $\mathrm{F}_\mathrm{s}\in\mathbf
{R}^{C_s^i\times T_s\times H_s^i\times W_s^i}$ and $\mathrm{F}_\mathrm{d}\in\mathbf{R}^{C_d^i\times T_d\times H_d^i\times W_d^i}$ from $S$ and $D$, respectively.

\subsection{Multi-scale Cross-level Fusion 
 (MCF) Module}
\begin{figure}[!t]
\centering
\includegraphics[width=0.48\textwidth]{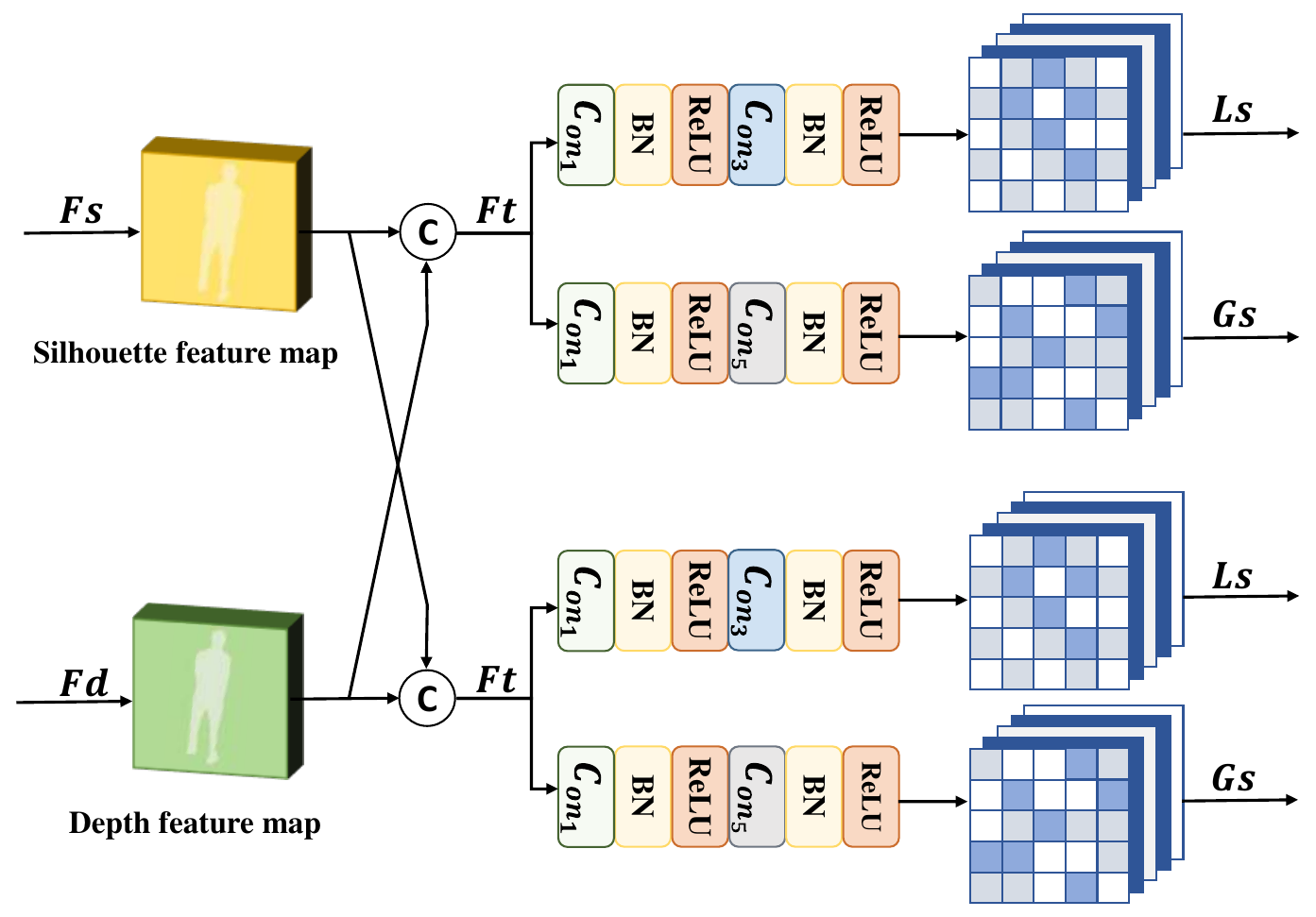} 
\caption{The detailed structure of MSSE.}
\label{fig5}
\end{figure}

The multi-scale spatial extraction component (MSSE) begins by concatenating the feature maps $F_s$ and $F_d$ along the channel dimension (see Figure~\ref{fig5}), resulting in a unified feature tensor $F_t$:
\begin{equation}
F_t = \text{Concat}(F_s, F_d) \in \mathbf{R}^{C_t^i \times T_t \times H_t^i \times W_t^i}.
\end{equation}
This concatenation enables subsequent convolution operations to process these features simultaneously, leveraging the complementary information from both silhouette and depth modalities. 
Following the feature concatenation, the MSSE process is applied to the fused feature tensor $F_t$. This stage employs convolutional kernels of varying sizes to capture both fine-grained features and global information simultaneously. By utilizing this multi-scale approach, we enhance the comprehensiveness of the feature representation, ensuring that both local details and broader contextual information are preserved. 
 The local score $L_S$ and Global score $G_S$ can be obtained: 
\begin{align}
L_S &= \text{Con}_1(\Gamma(\text{Con}_3(\Gamma(\text{Con}_1(F_t))))), \\
G_S &= \text{Con}_1(\Gamma(\text{Con}_5(\Gamma(\text{Con}_1(F_t))))),
\end{align}
where $\Gamma(\cdot) = \text{BN}(\text{ReLU}(\cdot))$, $Con_1$  represents a 1×1 convolution kernel.
 Afterward, $L_S$ and $G_S$ are added to calculate the attention weight through M-ATF. Then, we fuse the feature maps $F_s$ and $F_d$ through the attention weight to obtain the fusion output of this stage $X_f$, allowing the model to selectively focus on important parts of the input, thereby improving the effectiveness of feature representation.  This process can be represented as:
\begin{align}
W_S &= \text{M-ATF}(\text{reshape}(L_S + G_S)), \\
X_f &= F_s \cdot W_S[:, 0] + F_d \cdot W_S[:, 1],
\end{align}
where M-ATF$(\cdot)$ represents Softmax$(\cdot)$.
To enhance the representation ability of features, we use a cross-level fusion method to obtain the final gait representation $Y_f$ at this stage, while obtaining higher-level semantic information and retaining rich fine-grained spatial information contained in the shallow feature map.
\begin{equation}
Y_f = X_f + F_{sd},
\end{equation}
where $F_{sd}$ represents the encoder's output at each stage, similar to $F_{s}$ and $F_{d}$. DeepGaitV2 is a multi-stage feature extraction network, in which encoding and fusion operations are performed at each stage. To keep the pipeline easy to read, we only show one stage in Figure~\ref{fig2}. After the first-stage fusion, the features input to both SFE and DFE become identical. Therefore, we use $F_{sd}$ to represent these unified features. Since Fsd is generated after fusion, it is represented by a dashed line.
To address the issue of insufficient information interaction between modalities, we implement the fusion process at each stage of DeepGaitV2. This stage-by-stage fusion strategy integrates feature information from different depth layer by layer, progressively optimizing feature representation and gradually combining information to enhance feature expression capability and improve overall model performance.

\subsection{Prediction Module}
In this module, we follow the traditional gait recognition framework. First, we apply temporal pooling (TP) ~\cite{nguyen2018weakly} and horizontal pooling (HP) ~\cite{fu2019horizontal} to the feature $Y_f$.  Next, fully connected layers (FCs) and BNNecks are used to obtain $f$ and $p$, which are then utilized to calculate the triplet loss. In the training stage, our  framework is trained in an end-to-end manner, and a combined loss is defined as:
\begin{equation}
\mathcal{L} = \alpha \mathcal{L}_{\text{tri}} + \beta \mathcal{L}_{\text{ce}}
\end{equation}
where $L_{tri}$ is the triplet loss ~\cite{hermans2017defense}, $L_{ce}$ is the cross-entropy loss ~\cite{zhang2018generalized}, $\alpha$ and $\beta$ are weighting hyperparameters, respectively.

\setlength{\tabcolsep}{3mm}
\renewcommand{\arraystretch}{1.1} 

\begin{table}
\centering
\caption{The amount of the identities (ID) and sequences (Seq)
covered by the employed datasets.}
\resizebox{0.48\textwidth}{!}{ 
\begin{tabular}{c | c c c c c}
\toprule[1.5pt]
\multirow{2}{*}{Data Set} & \multicolumn{2}{c}{Train Set} & \multicolumn{2}{c}{Test Set} & \multirow{2}{*}{Scenes}\\ 
& ID & Seq & ID & Seq & \\ 
\cmidrule{1-6}
CCPG & 100 & 8388 & 100 & 8178 & CL, UP, DN, BG \\
SUSTech1K & 250 & 6011 & 750 & 19228 & Various \\
CASIA-B* & 74 & 8140 & 50 & 5500 & NM, BG, CL \\ 
\bottomrule[1.5pt]
\end{tabular}
}

\label{Table 1}
\end{table}

\setlength{\tabcolsep}{6pt} 
\renewcommand{\arraystretch}{1.0} 
\begin{table*}[ht]
\caption{Results of gait recognition and person re-identification of different models on the CCPG dataset. The best and second best performances are highlighted in bold and underlined, respectively.}
\centering
\resizebox{0.97\textwidth}{!}{
\begin{tabular}{c|c c|c c c c| c|c c c c| c}
\toprule[1.5pt]
\multirow{2}{*}{Input} & \multirow{2}{*}{Model} & \multirow{2}{*}{Venue} & \multicolumn{5}{c|}{Gait Evaluation Protocol} & \multicolumn{5}{c}{ReID Evaluation Protocol} \\ \cmidrule{4-13}
& & & CL & UP & DN & BG & Mean & CL & UP & DN & BG & Mean \\ \midrule
\multirow{5}{*}{Silhouette} 
& GaitSet & AAAI'19 & 60.2 & 65.2 & 65.1 & 68.5 & 64.8 & 77.5 & 85.0 & 82.9 & 87.5 & 83.2 \\ 
& GaitPart & CVPR'20 & 64.3 & 67.8 & 68.6 & 71.7 & 68.1 & 79.2 & 85.3 & 86.5 & 88.0 & 84.8 \\ 
& AUG-OGBase & CVPR'23 & 52.1 & 57.3 & 60.1 & 63.3 & 58.2 & 70.2 & 76.9 & 80.4 & 83.4 & 77.7 \\ 
& GaitBase & CVPR'23 & 71.6 & 75.0 & 76.8 & 78.6 & 75.5 & 88.5 & 92.7 & \underline{93.4} & 93.2 & 92.0 \\ 
& DeepGaitV2 & Arxiv'23 & 78.6 & \underline{84.8} & 80.7 & 89.2 & 83.3 & \underline{90.5} & \underline{96.3} & 91.4 & 96.7 & 93.7 \\ \midrule
\multirow{3}{*}{Skeleton} 
& GaitGraph2 & CVPRW'22 & 5.0 & 5.3 & 5.8 & 6.2 & 5.1 & 5.0 & 5.7 & 7.3 & 8.8 & 6.7 \\ 
& Gait-TR & ES'23 & 15.7 & 18.3 & 18.5 & 17.5 & 17.5 & 24.3 & 28.7 & 31.1 & 28.1 & 28.1 \\ 
& MSGG & MTA'23 & 29.0 & 34.5 & 37.1 & 33.3 & 33.5 & 43.1 & 52.9 & 57.4 & 49.9 & 50.8 \\ \midrule
{Skeleton Map}& SkeletonGait & AAAI'24  & 40.4& 48.5 & 53.0 & 61.7 & 50.9 & 52.4 & 65.4 & 72.8 & 80.9 & 67.9 \\ \midrule
{Human Parsing}& DeepGaitV2 & Arxiv'23  & 69.6& 75.8& 75.8 & 83.3 & 76.1 & - & - & - & - & - \\ \midrule
 \rowcolor[HTML]{EAEAEA}
{Depth Map} & \textbf{DepthGait$\star$} & \textbf{Ours} & \underline{79.1} & 83.8 & \underline{84.3} & \underline{90.6} & \underline{84.5} & 90.3 & 94.5 & 93.1 & \underline{97.1} & \underline{93.8} \\ \midrule

\multirow{4}{*}{Multimodal} 
& BiFusion & MTA'24 & 62.6 & 67.6 & 66.3 & 66.0 & 65.6 & 77.5 & 84.8 & 84.8 & 82.9 & 82.5 \\ 
& ParsingGait & MM'23 & 55.3 & 58.9 & 64.0 & 66.7 & 61.2 & 73.5 & 78.4 & 85.2 & 87.0 & 81.0 \\ 
& XGait & MM'24 & 72.8 & 77.0 & 79.1 & 80.5 & 77.4 & - & - & - & - & - \\ 
& SkeletonGait++ & AAAI'24 & \underline{79.1} & 83.9 & 81.7 & 89.9 & 83.7 & 90.2 & 95.0 & 92.9 & 96.9 & \underline{93.8} \\ \midrule

\rowcolor[HTML]{EAEAEA}
{Sils+Depth Map}&  
\textbf{DepthGait} & \textbf{Ours} & \textbf{83.9} & \textbf{87.9} & \textbf{86.0} & \textbf{92.6} & \textbf{87.6} & \textbf{92.7} & \textbf{96.6} & \textbf{94.2} & \textbf{97.2} & \textbf{95.2} \\ 
\bottomrule[1.5pt]
\end{tabular}
}

\label{Table 2}
\end{table*}

\section{Experiments}

\subsection{Datasets} We evaluate our approach on CCPG~\cite{li2023depth}, SUSTech1K~\cite{shen2023lidargait} and CASIA-B*~\cite{liang2022gaitedge}. The key statistics of these gait datasets are listed in Table~\ref{Table 1}. We adhere to the official protocols, which include the partition strategies for the training and gallery/probe sets. For testing, all datasets utilize comprehensive gait evaluation protocols in multi-view scenarios.

\subsection{Implementation Details} For CCPG, the batch size is set to 8 $\times$ 16, where 8 represents the number of IDs, and 16 for the training sequences per ID. And the data sampler collects a fixed-length segment of 30 frames as input. The Milestones are set to 20k, 30k, and 40k. The total steps are 60k in our implementation. We utilize the SGD optimizer with an initial learning rate of 0.1 and weight decay of 0.0005. The input size of depth maps is 64 $\times$ 44. At the test phase, the entire sequence of depth maps will be directly fed into  DepthGait. For SUSTech1K, the batch size is set to 8 $\times$ 8. The learning rate is reduced by a factor of 0.1 at the 20k, 30k, and 40k iterations, and the total number of iterations is set to 50k. And the data sampler collects a fixed-length segment of 10 frames as input. For CASIA-B*, all settings are the same as CCPG, except Milestones are set to 20k, 40k, 50k.

\subsection{Comparison with State-of-the-Art Methods}
\textbf{Evaluation on CCPG.} We have conducted a comprehensive comparison with several gait recognition methods in six categories: For the silhouette-based approach, peer methods involved for comparison include GaitSet \cite{chao2019gaitset}, GaitPart \cite{fan2020gaitpart}, GaitBase \cite{Fan_2023_CVPR}, and DeepGaitV2 \cite{fan2024exploringdeepmodelspractical}; For the skeleton-based approach,  GaitGraph \cite{teepe2021gaitgraph}, Gait-TR \cite{zhang2023spatial}, MSGG \cite{10.1007/s11042-023-15483-x}. Specially, we also compared two new modalities, HumanParsing~\cite{10.1145/3581783.3612052}and SkeletonGait \cite{fan2024skeletongait}; For the multimodal approach, peer methods involved for comparison include BiFusion \cite{peng2024learning}, ParsingGait \cite{10.1145/3581783.3612052}, XGait\cite{zheng2024takes}, and SkeletonGait++ \cite{fan2024skeletongait}. To demonstrate the superiority of depth maps derived from RGB images in gait recognition, we also implement a variant of our model, termed DepthGait$\star$ (using depth image only), and compare it with peer monomodality-based methods.

\begin{figure}[!t]
\centering
\includegraphics[width=0.48\textwidth]{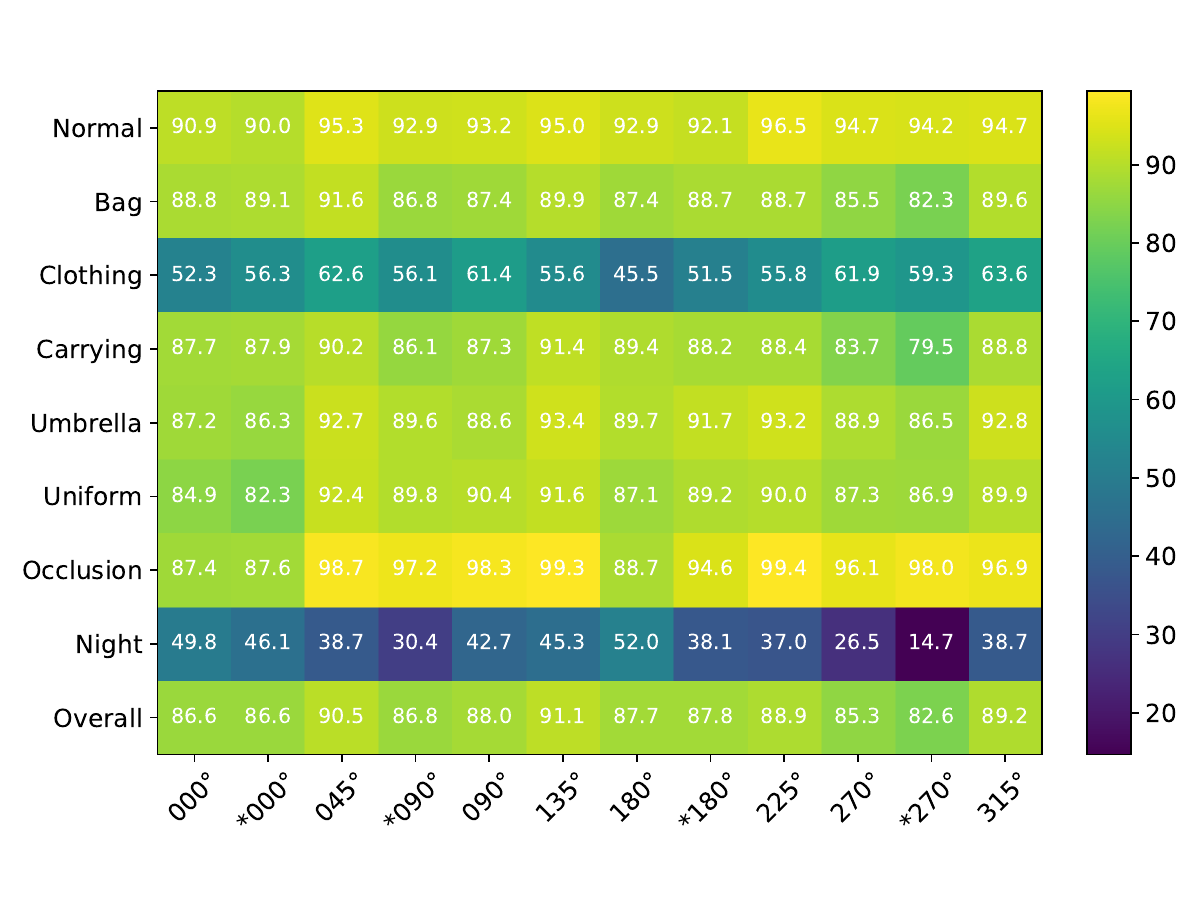} 
\caption{The performance of DepthGait in different attributes and views on the SUSTech1K dataset. *indicates viewpoint at a longer distance.}
\label{fig6}
\end{figure}

As can be observed in Table~\ref{Table 2}, the proposed DepthGait$\star$, which uses only depth maps as input, outperforms all peer methods with a mean rank-1 accuracy of 84.5\% in gait recognition task and 93.8\% in person re-identification task, respectively, and our multimodal DepthGait improves gait recognition performance further, achieving a mean rank-1 accuracy of 87.6\% in gait recognition task and 95.2\% in person re-identification task, respectively.
 Notably, recognition accuracy significantly increases in the CL, UP, and DN conditions. 
 Depth maps are less sensitive to such appearance changes, focusing more on the geometric structure of the object rather than surface texture, resulting in higher recognition accuracy. The superior performance of DepthGait across all modalities can be attributed to the depth map's ability to capture distance information from each point in the scene to the camera, providing additional 3D structural information that aids in detecting more detailed changes in body shape.

\begin{table*}[!t]
\centering
\caption{Evaluation on the SUSTech1K dataset.}
\setlength{\tabcolsep}{2pt}
\renewcommand{\arraystretch}{1.0} 
\resizebox{0.97\textwidth}{!}{
\begin{tabular}{c|c c|c c c c c c c c| c|c}
\toprule[1.5pt]
\multirow{2}{*}{Input} & \multirow{2}{*}{Model} & \multirow{2}{*}{Venue} & \multicolumn{8}{c|}{Probe Sequence (R-1)} & \multicolumn{2}{c}{Overall} \\ \cmidrule{4-13}
& & & Normal & Bag & Clothing & Carrying & Umbrella & Uniform & Occlusion & Night & R-1 & R-5 \\ \midrule
\multirow{5}{*}{Silhouette} 
& GaitSet & AAAI'19 & 69.1 & 68.2 & 37.4 & 65.0 & 63.1 & 61.0 & 67.2 & 23.0 & 65.0 & 84.8 \\ 
& GaitPart & CVPR'20 & 62.2 & 62.8 & 33.1 & 59.5 & 57.2 & 54.8 & 57.2 & 21.7 & 59.2 & 80.8 \\ 
& GaitGL & ICCV'21 & 67.1 & 66.2 & 35.9 & 63.3 & 61.6 & 58.1 & 66.6 & 17.9 & 63.1 & 82.8 \\ 
& GaitBase & CVPR'23 & 81.5 & 77.5 & \underline{49.6} & 75.8 & 75.5 & 76.7 & 81.4 & 25.9 & 76.1 & 89.4 \\ 
& DeepGaitV2 & Arxiv'23 & \underline{86.5} & 82.8 & 49.2 & 80.4 & \underline{83.3} & 81.9 & 86.0& 28.0 & 80.9 & 91.9 \\ \midrule
\multirow{3}{*}{Skeleton} 
& GaitGraph2 & CVPRW'22 & 22.2 & 18.2 & 6.8 & 18.6 & 13.4 & 19.2 & 27.3 & 16.4 & 18.6 & 40.2 \\ 
& Gait-TR & ES'23 & 33.3 & 31.5 & 21.0 & 30.4 & 22.7 & 34.6 & 44.9 & 23.5 & 30.8 & 56.0 \\ 
& MSGG & MTA'23 & 67.1 & 66.2 & 35.9 & 63.3 & 61.6 & 58.1 & 66.6 & 17.9 & 33.8 & - \\ \midrule
{Human Parsing}& DeepGaitV2 & Arxiv'23  & 84.7 & 77.3 & 29.4 & 78.2 & 75.8 & 80.2 & 87.9 & 43.3 & 77.3 & 91.3 \\ \midrule
{Skeleton Map}& SkeletonGait & AAAI'24  & 55.0 & 51.0 & 24.7 & 49.9 & 42.3 & 52.0 & 62.8 & 43.9 & 50.1 & 72.6 \\ \midrule
\multirow{4}{*}{Multi-modal} 
& BiFusion & MTA'24 & 69.8 & 62.3 & 45.4 & 60.9 & 54.3 & 63.5 & 77.8 & 33.7 & 62.1 & 83.4 \\ 
& ParsingGait & MM'23 & 82.7 & 73.3 & 30.4 & 76.4 & 70.8 & 74.6 & 84.3 & \textbf{50.5} & 75.0 & 90.7 \\ 
& SkeletonGait++ & AAAI'24 & 85.1 & \underline{82.9} & 46.6 & \underline{81.9} & 80.8 & \underline{82.5} & \underline{86.2} & \underline{47.5} & \underline{81.3} & \textbf{95.5} \\ \midrule
\rowcolor[HTML]{EAEAEA} 
{Sils+Depth Map}& 
\textbf{DepthGait} & \textbf{Ours} & \textbf{93.5} & \textbf{88.0} & \textbf{56.8} & \textbf{87.4} & \textbf{90.0} & \textbf{88.5} & \textbf{95.2} & 38.4 & \textbf{87.6} & \underline{95.0} \\ 
\bottomrule[1.5pt]
\end{tabular}
}
\label{Table 3}
\end{table*}

\begin{table}[!t]

\centering
\caption{Evaluation on the CASIA-B* dataset.}
\renewcommand{\arraystretch}{1.2} 
\resizebox{0.48\textwidth}{!}{ 
\begin{tabular}{c |c c|c c c|c}
\toprule[1.7pt]
\multirow{2}{*}{Input} &\multirow{2}{*}{Method} &\multirow{2}{*}{Venue}& \multicolumn{4}{c}{Gait Recognition} \\ \cmidrule{4-7}
&&& NM & BG & CL & Mean\\ \midrule
\multirow{4}{*}{Silhouette} & GaitSet & AAAI'19 & 92.3 & 86.1 & 73.4 & 83.9\\ 
&GaitPart & CVPR'20 & 93.1 & 86.0 & 75.1 & 84.7\\
&GaitGL & ICCV'21 & 94.1 & 90.0 & \textbf{81.4} & 88.5\\
&GaitBase & CVPR'23 & \underline{96.5} & \underline{91.5} & 78.0 & \underline{88.7}\\ \midrule
\rowcolor[HTML]{EAEAEA} 
\textbf{Depth Map} & \textbf{DepthGait} & \textbf{ours} & \textbf{98.6} & \textbf{95.0} & \underline{80.1} & \textbf{91.2}\\
\bottomrule[1.7pt]
\end{tabular}
}
\label{Table 4}
\end{table}
\begin{table}[t]
\centering
\caption{The role of key modules in multi-scale and cross-level fusion. Exclude means excluding identical-view cases, and Include means including identical-view cases.}
\begin{tabular}{c c c c c c}
\toprule[1.5pt]
\multirow{2}{*}{Baseline} & \multirow{2}{*}{Depth Map} & \multirow{2}{*}{MSF} & \multirow{2}{*}{CLF} & \multicolumn{2}{c}{Rank-1 (mean)} \\ 
\cmidrule(lr){5-6}
& & & & Exclude & Include \\ 
\midrule
\cmark & & & & 83.3 & 84.8 \\
 & \cmark & & & 84.5 & 85.8 \\
\cmark & \cmark & \cmark & & 86.8 & 87.9 \\
\cmark & \cmark & \cmark & \cmark & \textbf{87.6} & \textbf{88.4} \\
\bottomrule[1.5pt]
\end{tabular}
\label{Table 5}
\end{table}

\textbf{Evaluation on SUSTech1K.} 
The evaluation results for the SUS-Tech1K Dataset are presented in Table \ref{Table 3}. Similar to the CCPG dataset, we categorized the methods into Silhouette-based, Skeleton-based, new modality and Multi-modal for comparison. The findings from Table \ref{Table 3} indicate that across nearly all metrics, Depth Map-based methods outperform other modality gait recognition approaches. On average, the rank-1 accuracy is improved by 6.3\% in each category, with an overall rank-1 accuracy reaching 87.6\% and a rank-5 accuracy of 95.0\%.

It is worth noting that our results do not perform well under night conditions. We speculate that the reason is that the quality of the silhouettes under night conditions is too poor, which interferes with the shape of the depth maps when making the mask, resulting in poor results. As shown in Figure~\ref{fig6}, performance is most unstable under Clothing and Night conditions, with large fluctuations across different viewing angles, suggesting these attributes are highly sensitive to changes in viewing angles. The model performs best at 45°, 135°, 225°, and 315°, which are typically located on the side of the person, providing more feature information and thus improving recognition accuracy. In contrast, the 0°, 90°, 180°, and 270°, corresponding to the front, back, and sides of the person, show relatively poor performance.

\textbf{Evaluation on CASIA-B*.} The evaluation results for the CASIA-B* Dataset are presented in Table \ref{Table 4}. To demonstrate the generalizability of the depth map modality across different datasets, we conducted the following experiments on the CASIA-B* dataset: First, we generated depth images from the RGB images in CASIA-B* according to the depth map generation method described earlier. We observed an improvement in rank-1 accuracy across the NM, BG, and CL scenarios, with an average increase of 2.5\%. The overall rank-1 accuracy reached 91.2\%, marking the highest performance among all gait recognition methods.

\subsection{Ablation Study} 

To demonstrate the effectiveness of our proposed DepthGait, we conducted a comprehensive ablation study on the CCPG dataset, as presented in Table \ref{Table 5}. As a baseline, we used only the silhouette sequence input with DeepGaitV2 as the feature extraction network for gait recognition. The initial mean rank-1 accuracy was 83.3\% in the exclude identical-view cases and 84.8\% in the include identical-view cases. When the input modality was switched from silhouette sequences to processed depth maps (indicated as Depth Map), the recognition accuracy improved by 1.2\% and 1.0\%, respectively. Introducing the MSF (Multi-scale Fusion) scheme, which fuses the depth map and silhouette sequence, further improved the accuracy by 3.5\% and 3.1\%. The CLF (Cross-Level Fusion) approach, which enhances feature integration at different levels, resulted in an additional improvement of 4.3\% and 3.6\%, leading to final recognition accuracies of 87.6\% and 88.4\%. These experimental results demonstrate that each component of our proposed method contributes to improving gait recognition accuracy. Below are more results of ablation experiments.

\textbf{Effectiveness of the Multi-scale Cross-level Fusion (MCF) Module.} As shown in Table \ref{Table 6}, we compared our method with several existing fusion techniques, including PlusFusion, CatFusion, Attention Fusion, and other multimodal fusion models. The results reveal that our MCF method outperforms all other fusion methods, contributing to overall improvements in CL, UP, DN, and BG by 4.8\%, 4.1\%, 1.7\%, and 2.0\%, respectively, compared to using only the depth map. These results demonstrate the effectiveness of our multi-scale, cross-level fusion module.

\textbf{The Impact of Different Channel Numbers on Gait Recognition.}
Varying the number of channels produces diverse feature maps, impacting gait recognition accuracy. A higher channel number results in a more complex model, which demands a more sophisticated feature fusion method. As previously discussed, the simplicity of the fusion method used in SMPLGait significantly diminishes the 3D branch's contribution when paired with an enhanced feature extraction network. Similarly, the proposed DepthGait utilizes a deeper feature extraction network, necessitating a more sophisticated fusion method. 

Table \ref{Table 7} illustrates the impact of different channel numbers on the overall fusion efficacy. It can be observed from the table that PlusFusion, CatFusion, and Attention Fusion are relatively insensitive to changes in channel number, with recognition accuracy fluctuating by approximately 0.5\%. However, our MCF method showed a variation of nearly 2.0\%, four times greater than conventional fusion methods. This highlights the necessity for a more advanced feature fusion technique.

\setlength{\tabcolsep}{1mm}
\renewcommand{\arraystretch}{1.0}
\begin{table}[!t]
\caption{Evaluation with different attributes of different fusion methods on the CCPG dataset. The best performance for each protocol is highlighted in bold.}
\centering
\begin{tabular}{c|c|c c c c |c}
\toprule[1.5pt]
\multirow{2}{*}{Input}&\multirow{2}{*}{\makecell{Fusion \\Method}} & \multicolumn{5}{c}{Gait Recognition} \\ \cmidrule{3-7}
&& CL & UP & DN & BG & Mean\\ \hline 
Depth& - & 79.1 & 83.8 & 84.3 & 90.6 & 84.5 \\ \midrule

\multirow{3}{*}{\makecell{Sils+\\Others}} &BiFusion & 62.6 & 67.6 & 66.3 & 66.0 & 65.6 \\ 
&ParsingGait & 55.3 & 58.9 & 64.0 & 66.7 & 61.2\\ 
&SkeletonGait++ & 79.1 & 83.9 & 81.7 & 89.9 & 83.7 \\ \midrule
\multirow{4}{*}{\makecell{Depth +\\ Silhouette}} & PlusFusion & 81.6 & 86.0 & 85.1 & 91.7 & 86.1\\ 
&Cat Fusion & 82.0 & 86.6 & 85.3 & 92.1 & 86.5\\ 
&Attention Fusion & 81.6 & 86.3 & 85.3 & 91.6 & 86.2\\ \cmidrule{2-7}
\rowcolor[HTML]{EAEAEA} 
\textbf{MCF (Ours)} & \textbf{83.9} & \textbf{87.9} & \textbf{86.0} & \textbf{92.6} & \textbf{87.6}\\ 
\bottomrule[1.5pt]
\end{tabular}
\label{Table 6}
\end{table}

\begin{table}[!t]
\centering
\caption{The impact of different feature channel numbers on gait recognition on the CCPG dataset. The best performance for each protocol is highlighted in bold.}
\begin{tabular}{c| c| c c c c |c }
\toprule[1.5pt]
\multirow{2}{*}{\makecell{Fusion \\Method}}&\multirow{2}{*}{\makecell{Channel \\Number}} & \multicolumn{5}{c}{Gait Recognition} \\ \cmidrule{3-7}
&& CL & UP & DN & BG & Mean\\ \midrule
\multirow{2}{*}{PlusFusion}&{c=32}& 80.6 & 85.5 & 85.6 & 90.6 & 85.6\\  \cmidrule{2-7}
&{c=64}& 81.6 & 86.0 & 85.1 & 91.7 & 86.1\\  \cmidrule{1-7}
\multirow{2}{*}{Cat Fusion}&{c=32}& 81.6 & 85.9 & 85.6 & 90.5 & 85.9\\  \cmidrule{2-7}
&{c=64}& 82.0 & 86.6 & 85.3 & 92.1 & 86.5\\  \cmidrule{1-7}
\multirow{2}{*}{Attention Fusion}&{c=32}& 80.7 & 85.4 & 85.9 & 90.8 & 85.7\\  \cmidrule{2-7}
&{c=64}& 81.6 & 86.3 & 85.3 & 91.6 & 86.2\\ \cmidrule{1-7}
\multirow{2}{*}{\textbf{MCF (Ours)}} &{c=32}& 81.3 & 85.7 & 85.3 & 91.1 & 85.9\\ \cmidrule{2-7}
 \rowcolor[HTML]{EAEAEA}
 {c=64}& \textbf{83.9} & \textbf{87.9} & \textbf{86.0} & \textbf{92.6} & \textbf{87.6}\\ 
\bottomrule[1.5pt]
\end{tabular}
\label{Table 7}
\end{table}

\setlength{\tabcolsep}{1mm}
\renewcommand{\arraystretch}{1.0} 
\begin{table}[!t]
\caption{The impact of different fusion stages on gait recognition on the CCPG dataset. The best performance for each protocol is highlighted in bold.}
\centering
\begin{tabular}{c| c|c c c c |c}
\toprule[1.5pt]
\multirow{2}{*}{\makecell{Fusion \\Method}}&\multirow{2}{*}{\makecell{Fusion \\ Stage}}& \multicolumn{5}{c}{Gait Recognition} \\ \cmidrule{3-7}
&& CL & UP & DN & BG & Mean\\ \midrule
\multirow{3}{*}{PlusFusion}&{Stage=2}& 81.6 & 86.0 & 85.1 & 91.7 & 86.1\\  \cmidrule{2-7}
&{Stage=3}& 71.1 &77.0 & 77.6 & 85.6 & 77.8\\  \cmidrule{2-7}
&{Stage=4}& 75.0 & 80.5 & 81.5 & 87.6 & 85.6\\  \cmidrule{1-7}
\multirow{3}{*}{Cat Fusion}&{Stage=2}& 82.0 & 86.6 & 85.3 & 92.1 & 86.5\\  \cmidrule{2-7}
&{Stage=3}& 78.0 & 82.7 & 83.0 & 89.4 & 83.3\\  \cmidrule{2-7}
&{Stage=4}& 82.3 & 85.8 & 85.9 & 90.7 & 85.9\\  \cmidrule{1-7}
\multirow{3}{*}{Attention Fusion}&{Stage=2}& 81.6 & 86.3 & 85.3 & 91.6 & 86.2\\ \cmidrule{2-7}
&{Stage=3}& 76.1 & 79.8 & 82.0 & 86.7 & 81.2\\  \cmidrule{2-7}
&{Stage=4}& 73.2 & 78.3 & 77.5 & 85.2 & 78.6\\  \cmidrule{1-7}
 \rowcolor[HTML]{EAEAEA}
{\textbf{MCF (Ours)}} &\textbf{All Stages}& \textbf{83.9} & \textbf{87.9} & \textbf{86.0} & \textbf{92.6} & \textbf{87.6}\\ 
\bottomrule[1.5pt]
\end{tabular}
\label{Table 8}
\end{table}
\textbf{The Impact of Different Fusion Stages on Gait Recognition.} Given the differences between depth map and silhouette modalities, we aim to enhance information interaction between these modalities to bridge the modality gap. Since DeepGaitV2 employs a multi-stage encoding process, we explored the impact of different fusion stages on the overall fusion efficacy by comparing PlusFusion, CatFusion, and Attention Fusion at stages 2, 3, and 4, respectively. As shown in Table \ref{Table 8}, PlusFusion and CatFusion perform relatively well at stages 2 and 4, while the recognition accuracy of Attention Fusion decreases as the fusion stage is delayed. Additionally, regardless of the stage, our MCF multi-stage fusion approach consistently outperforms the single-stage fusion effects. This is because MCF integrates fusion at each encoding stage, enhancing information interaction between the two modalities throughout the process. Early stages capture low-level features such as edges and textures, while later stages capture high-level semantic features like actions and postures. Moreover, unlike the single-scale fusion of Attention Fusion, our multi-scale method simultaneously captures local detail features and global structural features. Local features help identify subtle individual gait differences, while global features capture the overall gait pattern. Such a multi-stage, multi-scale fusion strategy contributes to the performance gains of the proposed method.
In addition to the above experiments, we also use visualization to analyze why the depth map is significantly better than the contour map.

\section{Conclusion}
In this paper, we present DepthGait, a novel framework that utilizes RGB-derived depth maps and silhouettes for improved gait recognition. Depth maps, unlike traditional representations, preserve both shape and structural information while incorporating pixel-level distance data, enabling the capture of finer body shape variations. This detail enhances gait feature distinction and boosts recognition accuracy. To bridge the modality gap between depth maps and silhouettes, we introduce a cross-level, multi-scale attention fusion scheme. This approach enhances information exchange between modalities and captures both local and global features. Compared to peer work, our method achieves state-of-the-art performance on widely used datasets.

\begin{acks}
This work was supported by the National Natural Science Foundation of China (No. 62176223, 62302535, 62371305), and in part by the
Guangdong Basic and Applied Basic Research Foundation (2023A1515011639, 2024A1515030025).
\end{acks}

\bibliographystyle{ACM-Reference-Format}
\bibliography{sample-base}

\end{document}